\documentclass[10pt,twocolumn]{article}
\usepackage[utf8]{inputenc} 
\usepackage[T1]{fontenc}    

\usepackage{jmlr2e}

\usepackage[margin=1in]{geometry}

\usepackage{blindtext}
\usepackage{setspace}

\usepackage[]{algorithm2e}
\usepackage{listings}
\usepackage{xcolor}

\usepackage{url}            
\usepackage{booktabs}       
\usepackage{amsfonts}       
\usepackage{nicefrac}       
\usepackage{microtype}      

\usepackage{amsmath}

\usepackage{fancyvrb}  
\usepackage{graphicx}
\graphicspath{{./_static/}{./tikz/}}
\usepackage{natbib}

\ShortHeadings{WAX-ML: A Python library for machine learning and
  feedback loops on streaming data}{Emmanuel Sérié}

\firstpageno{1}

\usepackage{multicol}

\begin{document}

\title{WAX-ML: A Python library for machine learning and feedback loops on streaming data \\
  \vspace{.4in} \includegraphics[width=0.3\textwidth]{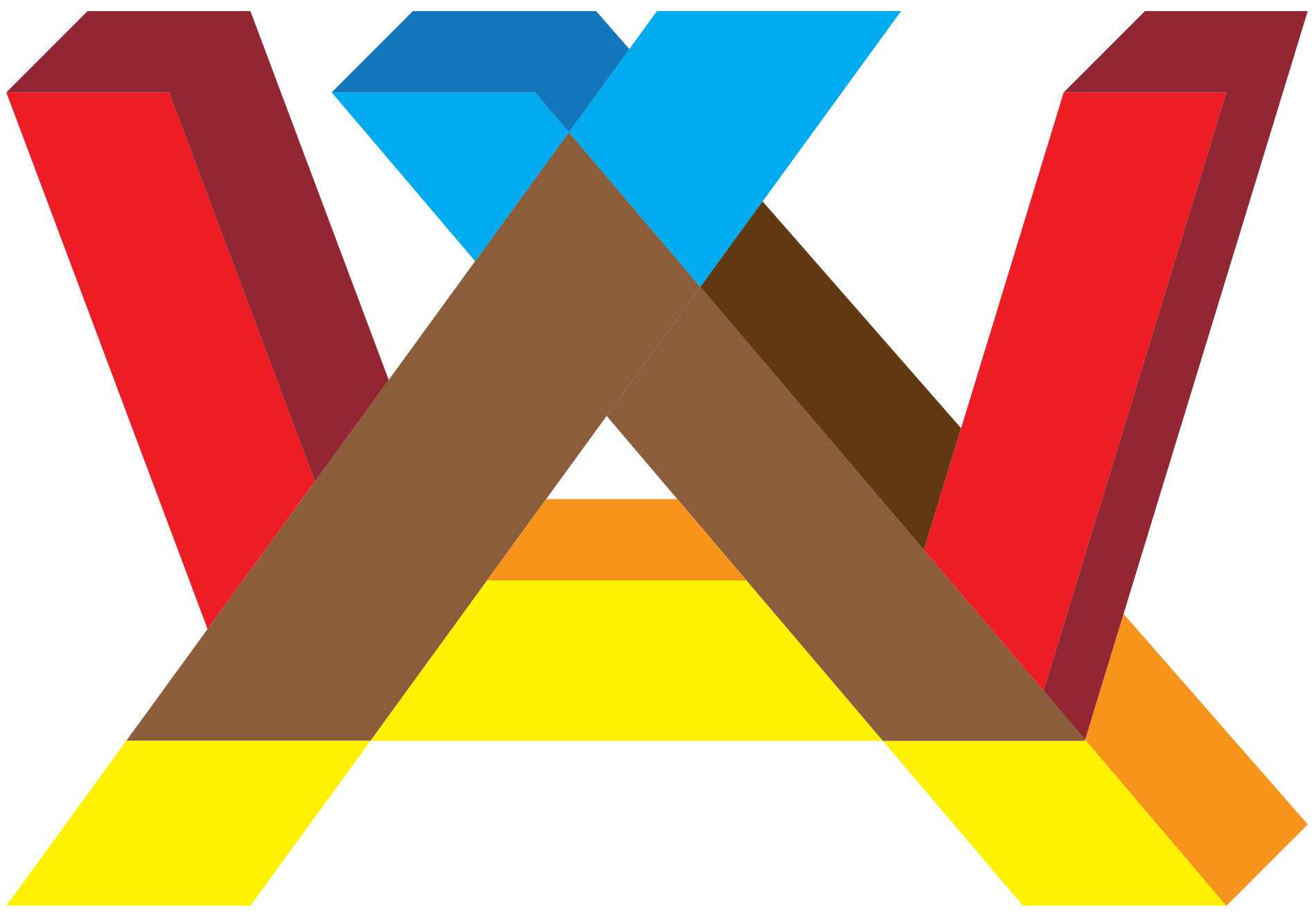}
  \vspace{.2in} 
  }

\author{ \name Emmanuel Sérié \email eserie@gmail.com \AND \addr
  Capital Fund Management \\ 23 rue de l'Université \\75007 Paris,
  France }

\twocolumn[
\begin{@twocolumnfalse}

  \maketitle

  \editor{}

  \begin{abstract}
    Wax is what you put on a surfboard to avoid slipping.  It is an essential tool to go surfing
    \dots We introduce WAX-ML a research-oriented Python library providing tools to design powerful
    machine learning algorithms and feedback loops working on streaming data.  It strives to
    complement JAX with tools dedicated to time series.  WAX-ML makes JAX-based programs easy to use
    for end-users working with pandas and xarray for data manipulation.  It provides a simple
    mechanism for implementing feedback loops, allows the implementation of online learning and
    reinforcement learning algorithms with functions, and makes them easy to integrate by end-users
    working with the object-oriented reinforcement learning framework from the Gym library.  It is
    released with an Apache open-source license on GitHub at \url{https://github.com/eserie/wax-ml}.
    \vspace{.1in}
  \end{abstract}

  \begin{keywords}
    machine-learning, streaming data, time series, feedback loops, Python, NumPy, JAX, Gym
    \vspace{.2in}
  \end{keywords}
\end{@twocolumnfalse}
]

\section{Introduction}

Since the advent of machine learning and its considerable development over the last 20 years many
tools and computer programs have been developed in this field.  The Python programming language
(see~\cite{python3}), which is a very flexible and easy-to-use programming language, has contributed
greatly to making machine learning programs more accessible.

Since its creation, a whole system of tools has been built to allow the implementation and access to
efficient numerical computation methods.  To begin with, the NumPy library, created in 2005 (see
\cite{harris2020array} for a recent review), allows to perform numerical computations and exposes,
with a simple API, multi-dimensional numerical arrays, and a set of functions implementing
mathematical operations on these arrays.

Subsequently, multiple tools have been built in different and complementary directions to go further
in the type of transformations that can be performed.

The Scikit-learn library, released in 2010 (see~\cite{sklearn_api}), has proposed a high-level
object-oriented API for exposing a very large set of machine learning methods for supervised and
unsupervised problems and has gathered a large international community around it.

On the other hand, the pandas library (see~\cite{mckinney-proc-scipy-2010}), launched in 2009 , has
built, on top of NumPy, a high-level powerful and flexible dataframe API that is better suited for
data analysis and manipulation tools.  In particular, it includes "time-series functionalities such
as date range generation and frequency conversion, moving window statistics, date shifting and
lagging, creation of domain-specific time offsets and joining of time-series".  It also strives to
have highly performance-optimized implementations with critical code paths written in Cython or C.

Some time-series oriented machine learning tools have been released, such as Prophet
(see~\cite{Taylor2017ForecastingAS}) in 2017 which is dedicated to end-users working on time series
and wanting to do prediction with black box statistical modeling techniques.

The TensorFlow library, released in 2015 (see~\cite{tensorflow2015-whitepaper}), proposed a set of
tools to implement deep neural networks and has since built a large community of developers and
users around it.  The TensorFlow developers rebuilt a tensor API in Python, aiming to be compatible
with NumPy but not completely, which made TensorFlow a separate ecosystem from that built around
NumPy.  In particular, this separation sometimes makes it difficult to develop research ideas in a
short development cycle, often requiring reimplementation of NumPy code in TensorFlow.  On the other
hand, its widespread adoption in the research community and industry for the implementation of
deep-learning algorithms has led to the development of very powerful optimization systems such as
XLA (Accelerated Linear Algebra) (see~\cite{xla}), a domain-specific linear algebra compiler that
can accelerate TensorFlow models with potentially no changes to the source code, resulting in
improvements in speed and memory usage (see~\cite{mlperf-bert}).

Since then, other libraries have been built to allow the implementation of XLA programs outside of
TensorFlow with other systems such as JAX (see~\cite{jax2018github}), Julia
(see~\cite{bezanson2017julia}), PyTorch (see~\cite{NEURIPS2019_9015}), Nx (see~\cite{Nx}).

In particular, JAX, a library released in 2018 (see~\cite{jax2018github}), has initiated a path to
solving the "ecosystem separation" problem explained above, by providing a functional
research-oriented API in Python for writing XLA programs with a NumPy API and primitive
transformations to perform just-in-time XLA compilation, automatic differentiation, parallelization,
and vectorization.

The approach followed by the JAX developers has been to propose some quite low-level primitives
working with well-identified programming constraints, such as working with pure functions.  This
constraint seems to be a good compromise between maintaining the flexibility of Python and having
the possibility to optimize numerical computations with systems like XLA.  This approach suggests
writing numerical Python programs in a functional programming style afterward.  This is not very
natural in a language like Python which is at its core an object-oriented programming language.
Once the constraints given by the functional programming approach of JAX are "digested", this can
become a strength since functional programming allows one to write notably modular, reusable,
testable programs, which aim at making JAX an efficient tool to implement research ideas.

This bet taken by JAX has somehow worked since a whole ecosystem has been built around it with the
advent of domain-specific libraries.  For instance, Optax (see~\cite{optax2020github}), "a gradient
processing and optimization library for JAX", Haiku (see~\cite{haiku2020github}), "a simple neural
network library for JAX", RLax (see~\cite{rlax2020github}), "a library built on top of JAX that
exposes useful building blocks for implementing reinforcement learning agents", Flax
(see~\cite{flax2020github}), "a neural network library and ecosystem for JAX designed for
flexibility", \dots

With the launch of WAX-ML, we want to complement this growing ecosystem with tools that make it
easier to implement machine learning algorithms on streaming data.

We want to push forward the efforts initiated by JAX developers to reconcile the Python ecosystems
around numerical computations by proposing tools allowing the use of JAX programs on time-series
data thanks to the use of the pandas and xarray (see~\cite{hoyer2017xarray}) libraries which are,
for now, reference libraries for the manipulation of labeled datasets and in particular time-series
data.

We even think that the TensorFlow and PyTorch ecosystem could also be made compatible with programs
implemented with JAX ecosystem tools thanks to EagerPy (see~\cite{rauber2020eagerpy}) which is a
library that allows writing code that works natively with PyTorch, TensorFlow, JAX, and NumPy.  In
this first released version of WAX-ML, we use EagerPy to facilitate the conversion between NumPy
tensors (which are used by pandas and xarray) and JAX tensors. We are also implementing some
universal modules so that we can work with other tensor libraries, but for now, we remain focused on
running JAX programs on streaming data.

\section{WAX-ML goal}

WAX-ML's goal is to expose "traditional" algorithms that are often difficult to find in standard
Python ecosystem and are related to time-series and more generally to streaming data.

It aims to make it easy to work with algorithms from very various computational domains such as
machine learning, online learning, reinforcement learning, optimal control, time-series analysis,
optimization, statistical modeling.

For now, WAX-ML focuses on time-series algorithms as this is one of the areas of machine learning
that lacks the most dedicated tools.  Working with time series is notoriously known to be difficult
and often requires very specific algorithms (statistical modeling, filtering, optimal control).

Even though some of the modern machine learning methods such as RNN, LSTM, or reinforcement learning
can do an excellent job on some specific time-series problems, most of the problems require using
more traditional algorithms such as linear and non-linear filters, FFT, the eigendecomposition of
matrices (\textit{e.g.}~\cite{bouchaud2005large}), principal component analysis (PCA) 
(\textit{e.g.}~\cite{DONG20181}), Riccati solvers for optimal control and filtering,\dots

By adopting a functional approach, inherited from JAX, WAX-ML aims to be an efficient tool to
combine modern machine learning approaches with more traditional ones.

Some work has been done in this direction in (\cite{leethorp2021fnet}) where transformer encoder
architectures are massively accelerated, with limited accuracy costs, by replacing the
self-attention sublayers with a standard, non-parameterized Fast Fourier Transform (FFT).

WAX-ML may also be useful for developing research ideas in areas such as online machine learning
(see~\cite{hazan_group, hazan2019introduction}) and development of control, reinforcement learning,
and online optimization methods.

\section{What does WAX-ML do?}

Well, building WAX-ML, we had some pretty ambitious design and implementation goals.

To do things right, we decided to start small and in an open-source design from the beginning.  We
released it with an Apache open-source license (see~\cite{wax-ml2021github}).

In this section, we give a quick overview of the features we have developed in WAX-ML.  They are
described in more detail in section~\ref{modules}.

For now, WAX-ML contains transformation tools that we call "unroll" transformations allowing us to
apply any transformation, possibly stateful, on sequential data.  It generalizes the RNN
architecture to any stateful transformation allowing the implementation of any kind of "filter".

We have implemented a "stream" module, described in section~\ref{synchro}, permitting us to
synchronize data streams with different time resolutions.

We have implemented some general pandas and xarray "accessors" permitting the application of any
JAX-functions on pandas and xarray data containers.

We have implemented a ready-to-use exponential moving average filter that we exposed with two APIs.
One for JAX users: as Haiku modules (see for instance our \verb+EWMA+ module).  A second one for
pandas and xarray users: with drop-in replacement of pandas \verb+ewm+ accessor.

We have implemented a simple module \verb+OnlineSupervisedLearner+ to implement online learning
algorithms for supervised machine learning problems.

We have implemented building blocks for designing feedback loops in reinforcement learning, and have
provided a module called \verb+GymFeedback+ allowing the implementation of feedback loop as the
introduced in the library Gym (see~\cite{gym}), and illustrated in Figure~\ref{fig:gym_feedback}.

Finally, we have implemented some ``universal'' modules that can work with TensorFlow, PyTorch, JAX,
and NumPy tensors.  At the moment, we have only implemented a demonstration module for the
exponential moving average that we have called \verb+EagerEWMA+.

\subsection{Why use WAX-ML?}

If you deal with time-series and are a pandas or xarray user, but you want to use the impressive
tools of the JAX ecosystem, then WAX-ML might be the right tool for you, as it implements pandas and
xarray accessors to apply JAX functions.

If you are a user of JAX, you may be interested in adding WAX-ML to your toolbox to address
time-series problems.

\subsection{Functional programming}

In WAX-ML, we pursue a functional programming approach inherited from JAX.

In this sense, WAX-ML is not a framework, as most object-oriented libraries offer.  Instead, we
implement "functions" that must be pure to exploit the JAX ecosystem.

We use the "module" mechanism proposed by the Haiku library to easily generate pure function pairs,
called \verb+init+ and \verb+apply+ in Haiku, to implement programs that require the management of
parameters and/or state variables.  In this way, we can recover all the advantages of
object-oriented programming but exposed in the functional programming approach.

This approach gives a lot of freedom in the type of ideas that can be implemented.  For instance,
JAX has been used recently to accelerate fluid dynamics simulations (see~\cite{kochkov2021machine})
by two orders of magnitude.

WAX-ML does not want to reinvent the wheel by reimplementing every algorithm.  We want existing
machine learning libraries to work well together while trying to leverage their strength, which is
easy to do with a functional programming approach.

To demonstrate this, in the current version of WAX-ML, we have constructed various examples, such as
an exponential moving average (to serve as a toy example), the implementation and calibration of an
LSTM architecture with a standard supervised machine learning workflow, and the implementation of
online learning and reinforcement learning architectures.  They are treated equally and laid out
with flat organization in our sub-package \verb+wax.modules+.

\subsection{Synchronize streams}
\label{synchro}

Physicists have brought a solution to the synchronization problem called the Poincaré–Einstein
synchronization (see~\cite{wiki:einstein_synchronisation}).  In WAX-ML we have implemented a similar
mechanism by defining a "local time", borrowing Henri Poincaré terminology, to denominate the
timestamps of the stream (the "local stream") in which the user wants to apply transformations and
unravel all other streams.  The other streams, which we have called "secondary streams", are pushed
back in the local stream using embedding maps which specify how to convert timestamps from a
secondary stream into timestamps in the local stream.

This synchronization mechanism permits to work with secondary streams having timestamps at
frequencies that can be lower or higher than the local stream. The data from these secondary streams
are represented in the "local stream" either with the use of a forward filling mechanism for lower
frequencies or with a buffering mechanism for higher frequencies.

Note that this simple synchronization scheme assumes that the different event streams have fixed
latencies.

We have implemented a "data tracing" mechanism to optimize access to out-of-sync streams.  This
mechanism works on in-memory data.  We perform the first pass on the data, without actually
accessing it, and determine the indices necessary to later access the data. Doing so we are vigilant
to not let any "future" information pass through and thus guaranty a data processing that respects
causality.

The buffering mechanism used in the case of higher frequencies works with a fixed buffer size (see
the WAX-ML module \verb+wax.modules.Buffer+) to allow the use of JAX / XLA optimizations and
efficient processing.

We give simple usage examples in our documentation and in working example shown in
Figure~\ref{fig:syncewma}.



\subsection{Working with streaming data}

WAX-ML may complement JAX ecosystem by adding support for streaming data.

To do this, we implement a unique data tracing mechanism that prepares for fast access to in-memory
data and allows the execution of JAX tractable functions such as \verb+jit+, \verb+grad+,
\verb+vmap+ or \verb+pmap+ (see section \ref{synchro}).

This mechanism is somewhat special in that it works with time-series data.

The \verb+wax.stream.Stream+ object implements this idea.  It uses Python generators to synchronize
multiple streaming data streams with potentially different temporal resolutions.  It works on
in-memory data stored in \verb+xarray.Dataset+.

To work with real streaming data, it should be possible to implement a buffer mechanism running on
any Python generator and to use the synchronization and data tracing mechanisms implemented in
WAX-ML to apply JAX transformations on batches of data stored in memory. We have not yet implemented
such a mechanism but put it on our enhancement proposal list.

\subsection{Adding support for time types in JAX}

At the moment \verb+datetime64+ and \verb+string_+ types are not supported in JAX.

WAX-ML add support for these NumPy types in JAX by implementing an encoding scheme for
\verb+datetime64+ relying on pairs of 32-bit integers similar to \verb+PRNGKey+ implemented in JAX
for pseudo-random number generation.  We also implement an encoding scheme for \verb+string_+
relying on \verb+LabelEncoder+ from the Scikit-learn library.

By providing these two encoding schemes, WAX-ML makes it easy to use JAX algorithms on data of these
types.

Currently, the types of time offsets supported by WAX-ML are quite limited and we would like to
collaborate with the pandas, xarray, and Astropy (see~\cite{astropy:2013}) teams to develop further
the time manipulation tools in WAX-ML.

\subsection{pandas and xarray accessors}

WAX-ML implements pandas and xarray accessors to ease the usage of machine-learning algorithms
implemented with JAX-functions within high-level data APIs such as \verb+DataFrame+ and
\verb+Series+ in pandas and \verb+Dataset+ and \verb+DataArray+ in xarray.

To load the accessors, users should have to simply run \verb+register_wax_accessors()+ function and
then use the "one-liner" syntax \verb+<data-container>.stream(…).apply(…)+

\subsection{Implemented modules}
\label{modules}

We have some modules (inherited from Haiku modules) ready to be used in \verb+wax.modules+.  They
can be considered as "building blocks" that can be reused to build more advanced programs to run on
streaming data.  We have some "fundamental" modules that are specific to time series management, the
`Buffer` module which implements the buffering mechanism, the `UpdateOnEvent` module which allows
one to "freeze" the computations of a program and to update them on some events in the "local
stream".

\begin{figure}[h]
  \begin{center}
    \includegraphics[width=0.5\textwidth]{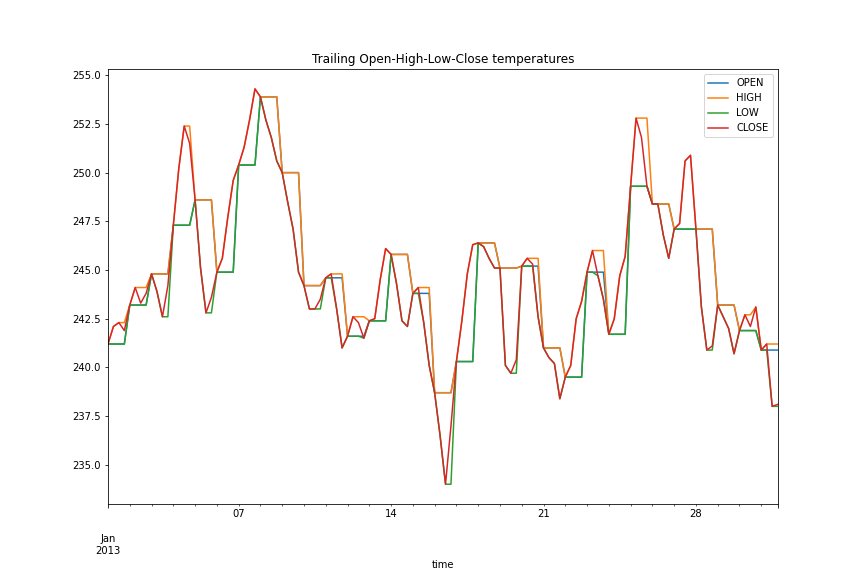}
  \end{center}
  \caption{Computation of trailing Open-High-Low-Close temperatures.}
  \label{fig:trailing_ohlc}
\end{figure}

To illustrate the use of this module, we show in Figure~\ref{fig:trailing_ohlc} how it can be used
to compute the trailing open, high, and close quantities of temperatures recorded during the day.

We have a few more specific modules that aim to reproduce some of the logic that pandas users may be
familiar with, such as \verb+Lag+ to implement a delay on the input data, \verb+Diff+ to compute
differences between values over time, \verb+PctChange+ to compute the relative difference between
values over time, \verb+RollingMean+ to compute the moving average over time., \verb+EWMA+,
\verb+EWMVar+, \verb+EWMCov+, to compute the exponential moving average, variance, and covariance of
the input data.

Finally, we implement domain-specific modules for online learning and reinforcement learning such as
\verb+OnlineSupervisedLearner+ (see section \ref{online}) and \verb+GymFeedback+ (see section
\ref{feedback}).

For now, WAX-ML offers direct access to some modules through specific accessors for pandas and
xarray users.  For instance, we have an implementation of the "exponential moving average" directly
accessible through the accessor \verb+<data-container>.ewm(...).mean()+ which provides a drop-in
replacement for the exponential moving average of pandas.

\begin{figure}[h]
  \begin{center}
    \includegraphics[width=0.5\textwidth]{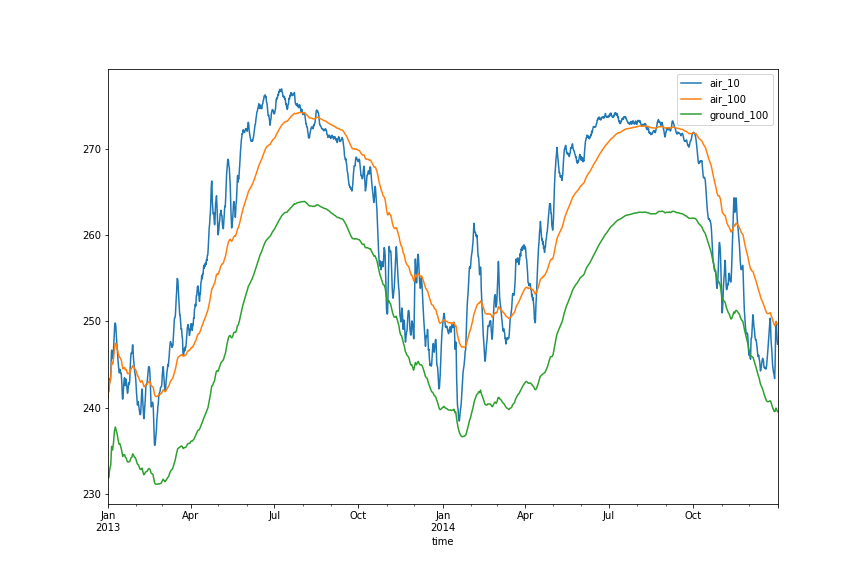}
  \end{center}
  \caption{We apply exponential moving average filters with different time scales
    while synchronizing two data streams ("air" and "ground")
    operating at different frequencies.}
  \label{fig:syncewma}
\end{figure}

We have implemented some working examples in our documentation.  In Figure~\ref{fig:syncewma}, we
show an illustrative example where we apply exponential moving average filters with different time
scales on a temperature data set (borrowed from the xarray library) while synchronizing two data
streams ("air" and "ground") operating at different frequencies.

\subsection{Speed}

The use of JAX allows for leveraging hardware accelerators that optimize programs for the CPU, GPU,
and TPU architectures.

We have performed few benchmarks with WAX-ML on the computation of exponential moving averages on
rather large dataframes (with 1 million rows and 1000 columns).  For this size of dataframe, we
observed a speedup of 3 compared to native pandas implementation.  This is not a very important
factor but it shows that even on such a simple algorithm already implemented in C, the XLA
compilation can optimize things.  The details of this benchmark are available in our documentation.

\subsection{Feedback loops}
\label{feedback}

Feedback is a fundamental notion in time-series analysis and has a wide history
(\textit{e.g.}~\cite{wiki:feedback}).  So, we believe it is important to be able to implement them
well in WAX-ML.

A fundamental piece in the implementation of feedback loops is the delay operator. We implement it
with the delay module \verb+Lag+ which is itself implemented with the \verb+Buffer+ module, a module
implementing the buffering mechanism.

The linear state-space models used to model linear time-invariant systems in signal theory are a
well-known place where feedbacks are used to implement for instance infinite impulse response
filters.  This is easily implemented with the WAX-ML tools and will be implemented at
a later time.

\begin{figure}[h]
  \begin{center}
    \includegraphics[width=0.5\textwidth]{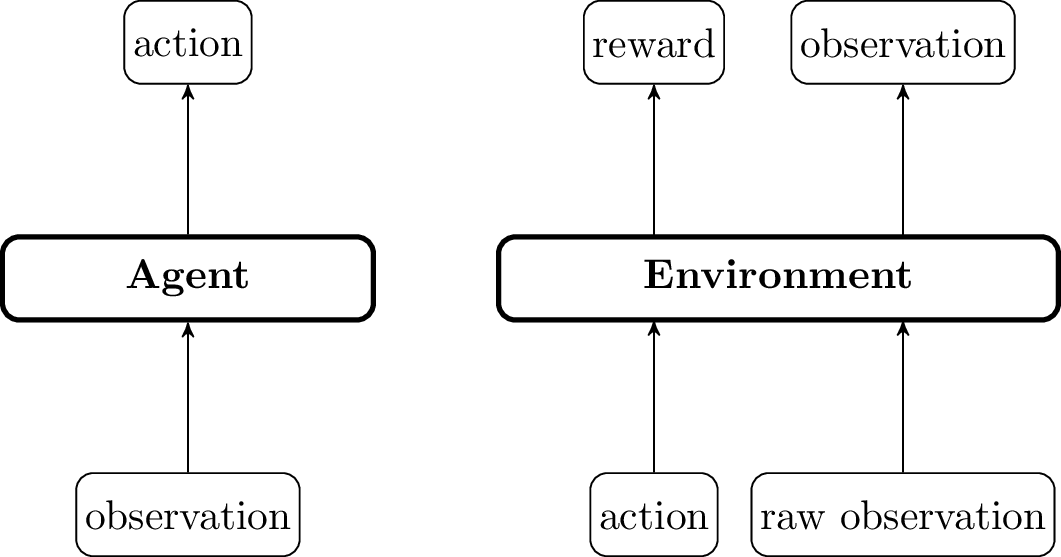}
  \end{center}
  \caption{Agent and Environment are simple functions (possibly with parameters and state).}
  \label{fig:agent_env}
\end{figure}

\begin{figure}[h]
  \begin{center}
    \includegraphics[width=0.5\textwidth]{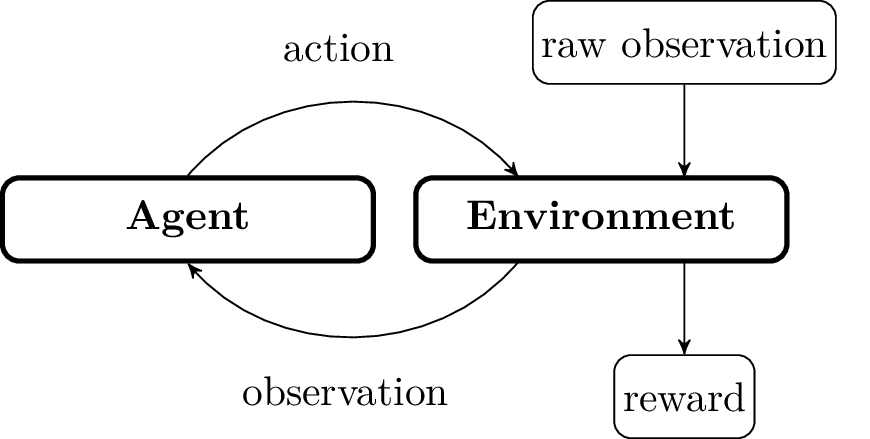}
  \end{center}
  \caption{Description of a Gym feedback loop.}
  \label{fig:gym_feedback}
\end{figure}

Another example is control theory or reinforcement learning. In reinforcement learning setup, an
agent and an environment interact with a feedback loop (see Figure~\ref{fig:gym_feedback}). This
generally results in a non-trivial global dynamic.  In WAX-ML, we propose a simple module called
\verb+GymFeedBack+ that allows the implementation of reinforcement learning experiments.  This is
built from an agent and an environment, both possibly having parameters and state (see
Figure~\ref{fig:agent_env}).  The agent is in charge of generating an action from observations.  The
environment is in charge of calculating a reward associated with the agent's action and preparing
the next observation from some "raw observations" and the agent's action, which it gives back to the
agent.

\begin{figure*}[h]
  \begin{center}
    \includegraphics[width=1\textwidth]{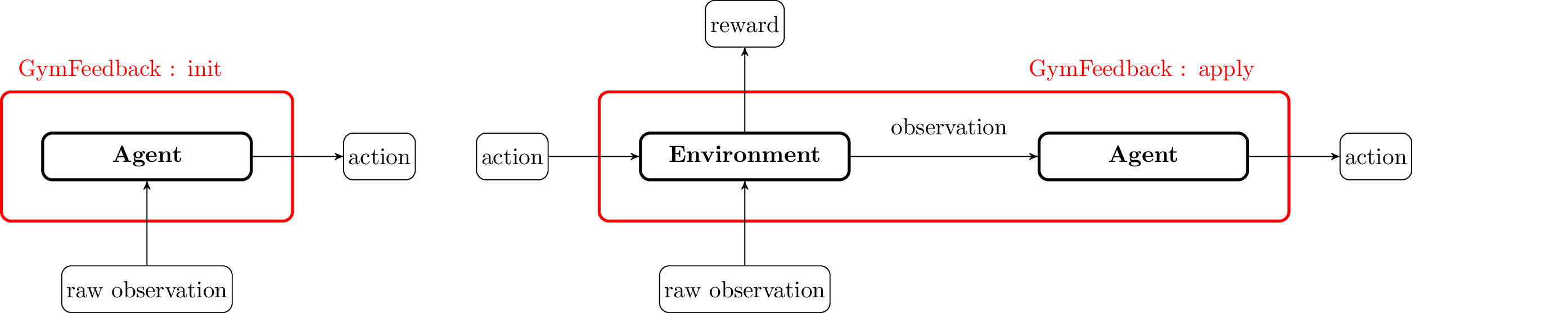}
  \end{center}
  \caption{Description of a Gym feedback loop in term of pure functions \texttt{init} and \texttt{apply}.}
  \label{fig:gym_feedback_init_apply}
\end{figure*}

A feedback instance \verb+GymFeedback(agent, env)+ is a function that processes the "raw
observations" and returns a reward as represented in the Figure~\ref{fig:gym_feedback}.
Equivalently, we can describe the function \verb+GymFeedback(agent, env)+, after transformation by
Haiku transformation, by a pair of pure functions \verb+init+ and \verb+apply+ that we describe in
Figure~\ref{fig:gym_feedback_init_apply}.

We have made concrete use of this feedback mechanism in the online learning application of
section~\ref{online}.

\subsection{Compatibility with other reinforcement learning frameworks}

In addition, to ensuring compatibility with other tools in the Gym ecosystem (see~\cite{gym}), we
propose a transformation mechanism to transform functions into standard stateful Python objects
following the Gym API for agents and environments as implemented in the \verb+deluca+ framework
(see~\cite{deluca}).

\section{Applications}
\subsection{Reconstructing the light curve of stars with LSTM}

\begin{figure}[h]
  \begin{center}
 \includegraphics[width=0.4\textwidth]{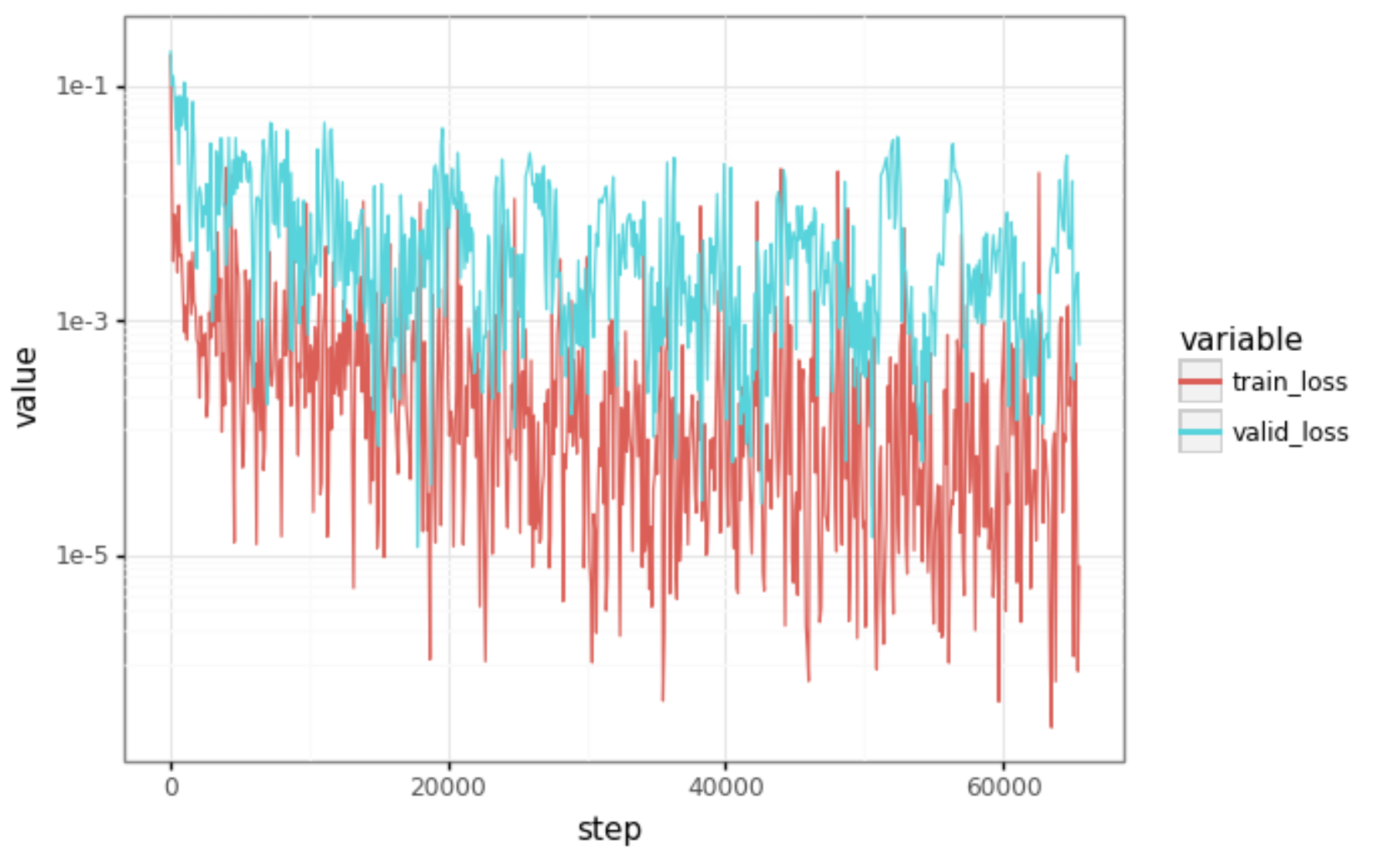}
 \includegraphics[width=0.4\textwidth]{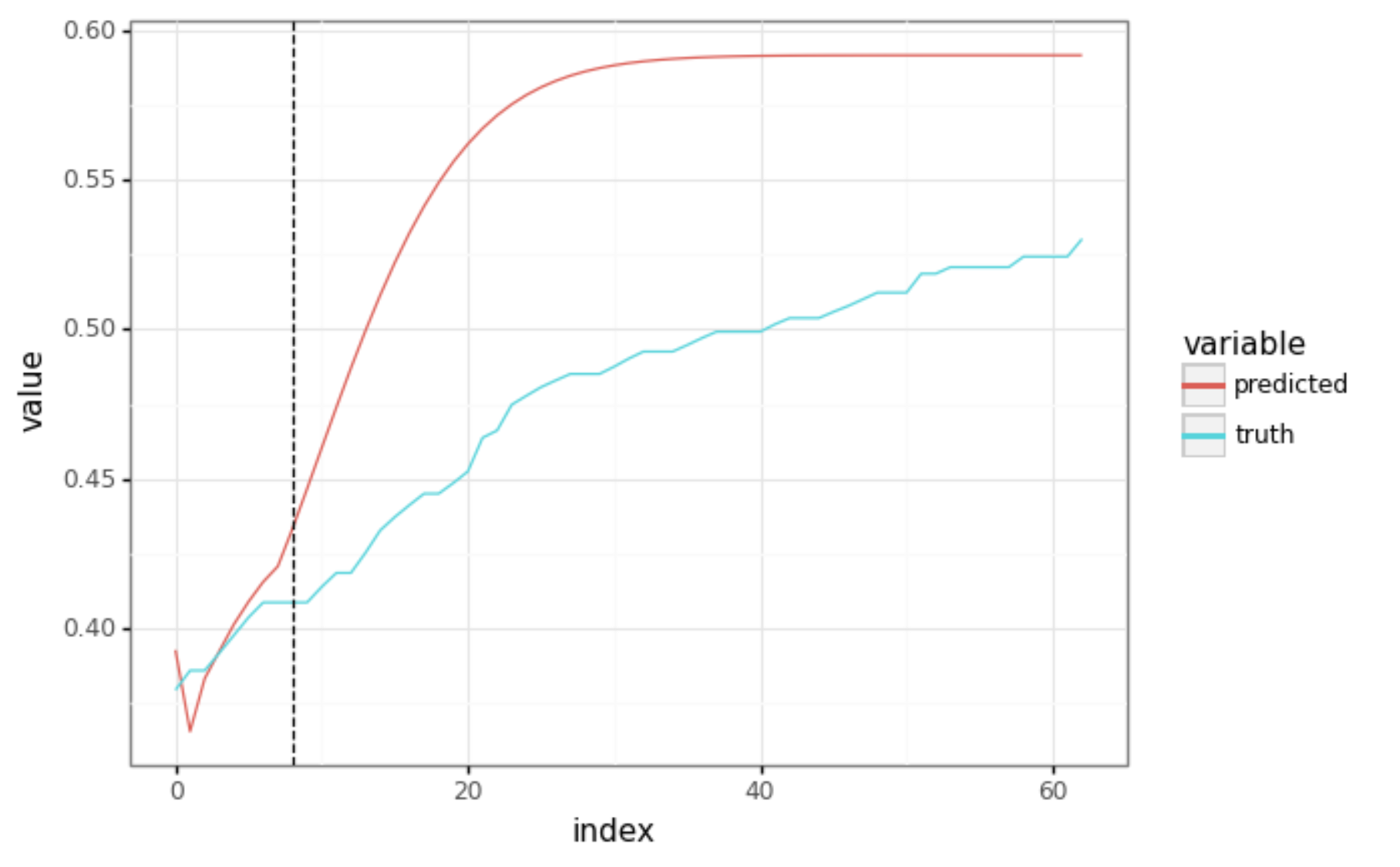}
\end{center}
\caption{light curve of stars reconstruction with LSTM. First: training curve.
  Second: auto-regressively predicted values and truth values.
}
\label{fig:lightcurve}
\end{figure}

To demonstrate more advanced usages of WAX-ML with machine learning tools on time series we have
reproduced the study (see~\cite{christophe_pere}) on the reconstruction of the light curve of stars.
We used the LSTM architecture to predict the observed light flux through time and use the Haiku
implementation.  We have then reproduced Haiku's case study for LSTM on this concrete application on
the light curve of stars.  We show an illustrative plot of the study in Figure~\ref{fig:lightcurve}.

Here WAX-ML is used to transition efficiently from time-series data with Nan values stored in a
dataframe to a "standard" deep learning workflow in Haiku.

The study is available in our documentation through a reproducible notebook.

As a disclaimer, we emphasize that although our study uses real data, the results presented in it
should not be taken as scientific knowledge, since neither the results nor the data source has been
peer-reviewed by astrophysists.  The purpose of the study was only to demonstrate how
WAX-ML can be used when applying a "standard" machine-learning workflow, here LSTM, to analyze time
series. However, we welcome further collaboration to continue the study.

\subsection{Online linear regression in non-stationary environment}
\label{online}

We performed a case study by implementing an online learning non-stationary linear regression
problem following online learning techniques exposed in (\cite{hazan2019introduction}).

We go there progressively by showing how a linear regression problem can be cast into an online
learning problem thanks to the \verb+OnlineSupervisedLearner+ module developed in WAX-ML.

Then, to tackle a non-stationary linear regression problem (i.e. with a weight that can vary in
time) we reformulate the problem into a reinforcement learning problem that we implement with the
\verb+GymFeedBack+ module of WAX-ML.

We then define an "agent" and an "environment" using simple functions implemented with modules as
illustrated in Figure~\ref{fig:agent_env}.  We have the agent which is responsible for learning the
weights of its internal linear model and the environment which is responsible for generating labels
and evaluating the agent's reward metric.  We finally assemble them in a gym feedback loop as in
Figure~\ref{fig:gym_feedback} and Figure~\ref{fig:gym_feedback_init_apply}.

\begin{figure*}[h]
  \begin{center}
\includegraphics[width=1.\textwidth]{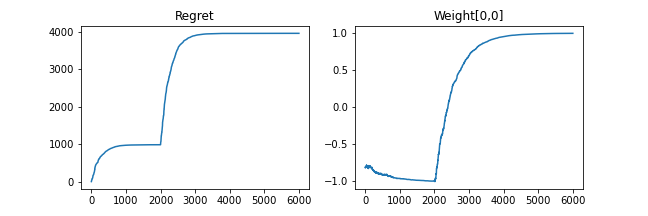}
\end{center}
\caption{Online linear regression in a non-stationary environment.  Left: The regret (cumulative sum
  of losses) first becomes concave, which means that the agent "learns something".  Then, the regret
  curve has a bump at step 2000 where it becomes locally linear.  It finally ends in a concave
  regime concave regime, which means that the agent has adapted to the new regime.  Right: We see
  that the weights converge to the correct values in both regimes.}
\label{fig:online}
\end{figure*}

We experiment with a non-stationary environment that flips the sign of the linear regression
parameters at a given time step, known only to the environment.  We show an illustrative plot of the
final result of the study in Figure~\ref{fig:online}.  More details about the study can be found in
our documentation\footnote{https://wax-ml.readthedocs.io/en/latest/}.

This example shows that it is quite simple to implement the online learning task with WAX-ML tools.
In particular, the functional programming approach allowed us to compose and to reuse functions to
design the final task.

\section{Future plans}

\subsection{Feedback loops and control theory}

We would like to implement other types of feedback loops in WAX-ML.  For instance, those of the
standard control theory toolboxes, such as those implemented in the SLICOT (see~\cite{Benner1999})
library.

Many algorithms in this space are absent from the Python ecosystem and we aim to provide JAX-based
implementations and expose them with a simple API.

An idiomatic example in this field is the Kalman filter, a now-standard algorithm that dates back to
the 1950s (see~\cite{Kalman_1960}).  After 30 years of existence, the Python ecosystem has still not
integrated this algorithm into widely adopted libraries.  Some implementations can be found in
Python libraries such as \verb+python-control+ (see~\cite{python-control}), statsmodels
(see~\cite{seabold2010statsmodels}), or SciPy (see~\cite{2020SciPy-NMeth}).  Also, some machine
learning libraries such as Scikit-learn (see~\cite{sklearn_api}) and River (see~\cite{2020river})
have some closed and non-solved issues on this subject.  Why has the Kalman filter not found its
place in these libraries?  We think it may be because they have an object-oriented API, which makes
them very well suited to the specific problems of modern machine learning but, on the other hand,
prevents them from accommodating additional features such as Kalman filtering.  We think the
functional approach of WAX-ML, inherited from JAX, could well help to integrate a Kalman filter
implementation in a machine learning ecosystem.

It turns out that Python code written with JAX is not very far from Fortran
(see~\cite{wiki:fortran}), a mathematical FORmula TRANslating system.  It should therefore be quite
easy and natural to reimplement standard algorithms implemented in Fortran, such as those in the
SLICOT library with JAX.  It seems that some questions about the integration of Fortran into JAX
have already been raised. As noted in an issue\footnote{issue 3950} reported on JAX's GitHub page,
and it might even be possible to simply wrap Fortran code in JAX, which would avoid a painful
rewriting process.

Along with the implementation of good old algorithms, we would like to implement more recent ones
from the online learning literature which somehow revisit the filtering and control problems.  In
particular, we would like to implement the online learning version of the ARMA model developed
in~\cite{pmlr-v30-Anava13} and some online-learning versions of control theory algorithms, an
approach called "the non-stochastic control problem", such as the linear quadratic regulator
developed in~\cite{hazan2020nonstochastic}.

\subsection{Optimization}

The JAX ecosystem already has a library dedicated to optimization: Optax, which we use in WAX-ML.
We could complete it by offering other first-order algorithms such as the Alternating Direction
Multiplier Method (ADMM) (\textit{e.g.}~\cite{BoydPCPE11}).  One can find functional implementations
of proximal algorithms in libraries such as \verb+proxmin+ (see~\cite{proxmin}),
\verb+ProximalOperators+ (see~\cite{proximaloperators}), or COSMO (see~\cite{garstka_2019}), which
could give good reference implementations to start the work.

Another type of work took place around automatic differentiation and optimization.  In
(\cite{agrawal2019differentiable}) the authors implement differentiable layers based on convex
optimization in the library \verb+cvxpylayers+. They have implemented a JAX API but, at the moment,
they cannot use the \verb+jit+ compilation of JAX yet. We would be interested in helping to solve
this issue\footnote{issue 103 JAX's GitHub page}.

Furthermore, in the recent paper by~\cite{blondel2021efficient}, the authors propose a new
efficient and modular implicit differentiation technique with a JAX-based implementation that should
lead to a new open-source optimization library in the JAX ecosystem.

 \subsection{Other algorithms}

 The machine learning libraries SciPy (see~\cite{2020SciPy-NMeth}), Scikit-learn
 (see~\cite{sklearn_api}), River (see~\cite{2020river}), and ml-numpy (see~\cite{mlnumpy}) implement
 many "traditional" machine learning algorithms that should provide an excellent basis for linking
 or reimplementing in JAX.  WAX-ML could help to build a repository for JAX versions of these
 algorithms.

 \section{Conclusion}

 We think that the simple tools we propose in WAX-ML might be useful for performing machine-learning
 on streaming data and implementing feedback loops.

 Some work could be continued to further develop the JAX ecosystem and implement some of the
 algorithms missing in the Python ecosystem in the wide adopted libraries.

 We hope that this software will be well received by the open-source community and that its
 development will also be actively pursued.

 \section{Collaborations}
 The WAX-ML team is open to discussion and collaboration with contributors from any field who are
 interested in using WAX-ML for their problems on streaming data.  We are looking for use cases
 around data streaming in audio processing, natural language processing, astrophysics, biology,
 finance, engineering ...

 We believe that good software design, especially in the scientific domain, requires practical use
 cases and that the more diversified these use cases are, the more the developed functionalities
 will be guaranteed to be well implemented.

 By making this software public, we hope to find enthusiasts who aim to develop WAX-ML further!

\section{Acknowledgements}

I thank my family for supporting me during this period of writing the first version of WAX-ML.

I thank my brother, Julien Sérié, musician, videographer, and talented graphic designer, who made
the logo of the software.

I would like to thank my colleagues at Capital Fund Management who encouraged me to pursue this
project and gave me constructive feedback while reading early drafts of this document.  I thanks
Richard Beneyton, Raphael Benichou, Jean-Philippe Bouchaud, David Eng, Tifenn Juguet, Julien Lafaye,
Laurent Laloux, Eric Lebigot, Marc Potters, Lamine Souiki, Marc Wouts, and the Open-source committee
of CFM for their early support. I thank Richard Beneyton, Gawain Bolton and Gilles Zerah for having
accepted to review the paper.

\bibliography{wax-ml-paper}
\end{document}